\documentclass{ifacconf}
\usepackage{natbib}   
\usepackage{placeins}
\usepackage{graphicx}     
\usepackage{amsmath}
\usepackage{tikz}
\usepackage{interval}
\usepackage{amssymb}
\usepackage{placeins}

\newcommand\figref{Fig.~\ref}
\newcommand\secref{Section~\ref}

\newcommand\tabref{Table~\ref}
\usepackage{booktabs}
\usepackage{amsmath}

\usepackage{todonotes}

\usepackage{xprintlen}

\usepackage{booktabs}
\usepackage[most]{tcolorbox}
\usepackage{multirow}
\usepackage{colortbl}

\usepackage{bm}
\usepackage{times}
\usepackage{microtype}

\newtheorem{definition}{Definition}

\makeatletter
\let\old@ssect\@ssect 
\makeatother

\usepackage{xcolor}
\definecolor{ccolor}{RGB}{0,19,114}

\usepackage{natbib}
\usepackage[hidelinks]{hyperref}
\hypersetup{
  colorlinks   = true, 
  urlcolor     = blue, 
  linkcolor    = blue, 
  citecolor   = ccolor 
}

\makeatletter
\def\@ssect#1#2#3#4#5#6{%
  \NR@gettitle{#6}
  \old@ssect{#1}{#2}{#3}{#4}{#5}{#6}
}
\makeatother

\begin{document}

\begin{frontmatter}

\title{Usage-Specific Survival Modeling Based on Operational Data and Neural Networks}

\author[First]{Olov Holmer} 
\author[First]{Mattias Krysander} 
\author[First]{Erik Frisk}

\address[First]{Department of Electrical Engineering, Linköping University,
SE-58183 Linköping, Sweden, 
   (e-mail: name.lastname@liu.se).}

\begin{abstract}  
Accurate predictions of when a component will fail are crucial when planning maintenance, and by modeling the distribution of these failure times, survival models have shown to be particularly useful in this context. Due to the complex behavior of degradation, describing this distribution is often a complex problem and data-driven methods are often preferred. In this paper, a methodology to develop data-driven
 usage-specific survival models that predict the distribution of the remaining life of
 a component based on its usage history is presented. 
 The methodology is based on conventional neural network-based survival models that are trained using data that is continuously gathered and stored at specific times, called snapshots. An important property of this type of training data is that it can contain more than one snapshot from a specific individual which results in that standard maximum likelihood training
 can not be directly applied since the data is not independent.
 However, the papers show that if the data is in a specific format
 where all snapshot times are the same for all individuals, called
 homogeneously sampled, maximum likelihood training can be
 applied and produce desirable results. In many cases, the data is not homogeneously sampled and
 in this case, it is proposed to resample the data to make it
 homogeneously sampled. How densely the dataset is sampled turns out to be an important parameter; it should be chosen large enough to produce good
 results, but this also increases the size of the dataset which
 makes training slow. To reduce the number of samples needed during training,
 the paper also proposes a technique to, instead of resampling the
 dataset once before the training starts, randomly resample the
 dataset at the start of each epoch during the training. The proposed methodology is evaluated on both a simulated dataset and an experimental dataset of starter battery failures. The results show that if the data is homogeneously sampled the methodology works as intended and produces accurate survival models. The results also show that randomly resampling the dataset on each epoch is an effective way to reduce the size of the training data and at the same improve the performance of the resulting model.

\end{abstract}


\end{frontmatter}

\section{Introduction}
In many applications, maintenance costs make up a significant portion of a system's total cost. To avoid unnecessary maintenance it is therefore of interest to determine the remaining useful life of the component so that maintenance can be planned as late as possible. In many cases, the degradation is not a deterministic process and predicting exactly when a component will fail is not possible, making a statistical description more useful. Survival models provide such a description by predicting the probability that the component will survive longer than a specific time. Due to the complex behavior of system degradation, such models are often difficult to find, and therefore data-driven methods are attractive.

While other methods like random survival forests \citep{ishwaranRandomSurvivalForests2008} exist,  neural network-based survival models have been shown to perform particularly well, and multiple models have been proposed for this, see for example \cite{brownUseArtificialNeural1997}, \cite{biganzoliFeedForwardNeural1998},
\cite{kvammeContinuousDiscretetimeSurvival2021}, and \cite{chingCoxnnetArtificialNeural2018}. However, these models are defined in a more general setting, and to use them for predicting the remaining life of a component is not always clear.

How a component is used often affects its useful life, and it is therefore of interest to base the predictions on operational data gathered as the component is used. Since degradation is often accumulative, the period during which the operational data is gathered can greatly affect the predictions. In \cite{holmer2023energy} this was solved by only considering operational data up to a specific age of the component, which means that the resulting model can only be used for predictions at this specific age. \cite{dhada2023weibull} used a similar approach but included usage data from gathered at multiple times during the components' lifetime; however, predictions were done based on a specific time and therefore only one prediction per component was made. In  \cite{liAttentionbasedDeepSurvival2022} the age of the component at the time of prediction was included as input to the model making it possible to use the model at any age. However, while the results from this wroke are impressive, the data used to train the model contains multiple data points from the same individuals and is therefore not independent, making standard training techniques not directly applicable, and it is not discussed why the method works or if it will work in other applications.

In conclusion, a more general definition of a usage-specific survival function that considers operational data gathered during any time interval is missing; as well as a methodology to train these types of models when there is more than one observation from each individual. The aim of this paper is therefore to address these two problems.

\section{Survival Modeling}\label{sec:survial_modeling}
 
Survival models describe the distribution of the failure time $T$, often conditioned on an explanatory variable $X$. They are often specified using the survival function, defined as 
\begin{equation} \label{eq:surv_function}
S(t; x) = P(T>t \mid X=x) = \int_t^\infty f(\tau ; x)\, d\tau
\end{equation} 
where $f$ is the corresponding failure probability density. In general, there are many density functions $f$ such that \eqref{eq:surv_function} holds, and the pair $(S,f)$ is thus needed to specify a survival model; however, in most cases, this is a more theoretical problem, and we will in most cases only use the survival function $S$ to specify a model.

\subsection{Survival Data and Censoring}\label{sec:conv_data}
In general, not all individuals are monitored up to the time they fail since, for example, the experiment ends or because of some other unconsidered failure in the system. This means that the data contains right-censoring and the data from $N$ individuals is on the form 
\begin{equation}\label{eq:conv_data}
  \mathcal{D} = \left\{ \left(\tau_i, \delta_i, x_i \right) \right\}_{i=1}^N
\end{equation}
where $\tau_i$ is the recorded time, $\delta_i$ is the indicator ($\delta_i=1$ for a failure, and $\delta_i=0$ for a censoring), and  $x_i$ is the covariate vector of individual $i$.

\subsection{Likelihood}

A likelihood function is defined based on a statistical model describing the distribution of the data. Since the model in our case only describes the distribution of a single observation it can not be used directly to define a likelihood function. However, by assuming that the observations are independent the model for the complete dataset factors into a product of the model for each observation. A consequence of this is that the likelihood function itself can be written as a product of the likelihood of each observation. This means that in this case defining a likelihood function based on independent observations is straightforward, as long as the likelihood of a single observation is known.  

Given a survival model parameterized by $\theta$, with survival function $S_\theta$ and density $f_\theta$, the likelihood of the observation $\left(\tau, \delta ,x \right)$ can be defined as
\begin{equation}
  L\left(\theta \mid \left(\tau, \delta ,x \right) \right) = 
  \begin{cases}
    f_\theta(\tau,x), & \delta=1\\
    S_\theta(\tau,x),& \delta =0
  \end{cases}.
\end{equation}
Assuming that the observations from each individual are independent, the likelihood of the dataset \eqref{eq:conv_data} factors into
\begin{equation}\label{eq:conv_lik}
L\left(\theta \mid \left(\tau, \delta ,x \right) \right) = 
\prod_{i:\delta_i=1} f_\theta\left(\tau_i, \delta_i, x_i\right) 
\prod_{i:\delta_i=0} S_\theta\left(\tau_i, \delta_i, x_i\right).
\end{equation}
In practice, the product above is numerically difficult to work with and instead the log-likelihood is used, defined as
\begin{equation}\label{eq:conv_loglik}
  \begin{split}
  l\left(\theta \mid \left(\tau, \delta ,x \right) \right) =& 
  \sum_{i:\delta_i=1} \log f_\theta\left(\tau_i, \delta_i, x_i\right)\\
  &+
  \sum_{i:\delta_i=0} \log S_\theta\left(\tau_i, \delta_i, x_i\right)  
  \end{split}.
  \end{equation}
\subsection{Neural Network-Based Survival Models and Maximum Likelihood Training}\label{sec:conv_training}
The dependencies on the explanatory variables are often complex making data-driven methods for predicting the survival function attractive, and in particular Neural network-based approaches have been shown to perform well. From \eqref{eq:conv_loglik} it follows that if a network parameterized by $\theta$ can provide the density $f_\theta$ and the survival function $S_\theta$ (along with their gradients), gradient-based learning can be applied to maximize the log-likelihood. In this context, for each sample $(\tau_i, \delta_i, x_i)$ the pair $(\tau_i,\delta_i)$ can be seen as the target and $x_i$ the features.

An important part of neural network-based survival modeling is to ensure that the pair $\left(S_\theta,f_\theta\right)$ defines a proper survival model; that is, the relation between $S_\theta$ and $f_\theta$ in \eqref{eq:surv_function} holds, and  $f_\theta(t,x)\geq0$. 
There are several ways to do this, see for example \cite{brownUseArtificialNeural1997}, \cite{biganzoliFeedForwardNeural1998}
\cite{kvammeContinuousDiscretetimeSurvival2021}, \cite{voronovPredictiveMaintenanceLeadacid2020}, \cite{chingCoxnnetArtificialNeural2018}, or \cite{liAttentionbasedDeepSurvival2022}.

In this work, the energy-based approach in \cite{holmer2023energy} is used. In this approach, a continuous survival function is defined by specifying the density function of $T$ as 
\begin{equation}
   f(t;x) = \frac{e^{-E_\theta(t;x)}}{Z(x)}, \qquad Z(x) = \int_0^\infty e^{-E_\theta(t;x)}\,dt 
\end{equation} 
where the energy $E_\theta(t;x)$ is taken as the output from a neural network, parameterized by $\theta$; and the normalization constant $Z(x)$ is evaluated numerically, for details on this see \cite{holmer2023energy}.  The survival function is then calculated as 
\begin{equation}
   \hat{S}_{\theta}(t;x) =1- \int_{0}^t f(\tau;x)\,d\tau = 1- \int_{0}^t \frac{e^{-E_\theta(\tau;x)}}{Z(x)}  \,d\tau
\end{equation}
where, again, the integral is evaluated numerically. Since the energy $E_\theta(t;x)$ is specified by a neural network, this model can essentially become as expressive as desired; however, this comes at the cost of the two numerical integrations. 

\section{Usage-Specific Survival Modeling Based on Operational Data}
We will now consider the problem of predicting the distribution of the failure time given the available operational data up to a specific age. The main difference compared with \secref{sec:survial_modeling} is that the explanatory variables now are based on aggregated system usage, which varies over time, and the main topic of the section is to define the usage-specific survival function and develop and train these type of models.

\subsection{Operational Data and Snapshots}
In many cases, how a system is used greatly affects the lifetime of the system, and information about this is useful for predicting when the system will fail. In this work, we assume that a snapshot $x(t_0)$ with information about how the system has been used up to age $t_0$ is available. The snapshot $x(t_0)$ could essentially contain any information about how the system has been used, and the only requirement is that it can be used as an input to a neural network. In practice, however, a trade-off between the amount of information stored in $x(t_0)$ and the amount of storage needed to store it must be done. For this reason, we in this work mainly consider snapshots on the form
\begin{equation}\label{eq:accumulative_ss}
  x(t_0) = \int_0^{t_0} g(y(t))\,dt
\end{equation}
where $y(t)$ denote the available measurements and control signals at time $t$, and $g$ is a function specifying how $y(t)$ is aggregated.

A simple example of a function $g$ is $g(y) = y$ which can be interpreted as accumulated usage; for example, if $y$ is the velocity of a truck then $x(t_0)$ would in this case be the total mileage up to age $t_0$. Another example is
\begin{equation}
    g(y) = 
    \begin{cases}
      1,&  y \leq y_{thres}\\
      0,&  y > y_{thres}
    \end{cases}
\end{equation}
where $y_{thres}$ is a threshold. In this case, $x(t_0)$ is the time spent with $y(t)\geq y_{thres}$ which, for example, can be used to measure the amount of time the power produced by an engine has been above a specific threshold. By using more than one threshold a histogram of how the component has been used can also be created; most of the features in the dataset in \secref{sec:battery_data} are created in this way.

Often data is continuously gathered and stored during the systems' lifetime, and consequently the available data from each individual is a sequence of snapshots $\left( x(t_i^{ss}) \right)_{i=1}^M$ where $t_1^{ss}<t_2^{ss}<\cdots<t_M^{ss}$ are the $M$ times when data was stored. However, when making a prediction, only one snapshot is used. It is of course likely that using by using a sequence of snapshots would result in better predictions; however, with the notation used in this work $x(t_i^{ss})$ should then be that sequence.  

\subsection{Usage-Specific Survival Function}\label{sec:us_surv_func}
A snapshot $x(t_0)$ describes how a specific individual has been used up age $t_0$. By interpreting $x(t_0)$ as an observation of a random variable $X(t_0)$, describing the distribution of the usage in a population, the usage-specific survival function can be defined as 
\begin{equation}
  S\left(t ; t_0, x \right) = P\left(T>t\mid X(t_0) = x\right),
\end{equation}
i.e., the probability of the failure happening after $t$ time units, given the operational data $x$ ($=x(t_0)$) at age $t_0$. 

Operational data often contain direct measurements from the considered component and, consequently, there is an implicit implication that $T>t_0$, since otherwise measurements from age $t_0$ would not be possible. If this is the case, it is more appropriate to consider the usage-specific survival function describing the remaining life of the component 
\begin{equation}\label{eq:rem_ub_surv}
  S^r\left(t ; t_0, x \right) = P\left(T>t+t_0\mid T>t_0,X(t_0) = x \right).
\end{equation}
However, to make the notation simpler, we will focus on presenting a methodology for predicting total life, and later in \secref{sec:remaining_life_modifications} show how to extend it to predicting remaining life.

\subsection{Survival Data Including Operational Data}
As in the conventional case in \secref{sec:conv_data}, the data from a specific individual contain a recorded time $\tau$ and an indicator $\delta$ indicating if it was a failure or censoring. However, instead of a single covariate vector $x$, we now have a sequence of samples $\left( x(t_i^{ss}) \right)_{i=1}^M$ from times $t_1^{ss}<t_2^{ss}<\cdots<t_M^{ss}$. The data from individual $i$ is therefore on the form
\begin{equation}
  \mathcal{D}_i = \left(\tau_i, \delta_i, \left\{\left(x_i(t_{i,j}^{ss}),t_{i,j}^{ss}\right)\right\}_{j=1}^{M_i}  \right)
\end{equation}
where $M_i$ is the number of snapshots from individual $i$, and for a population of $N$ individuals the data is on the form
\begin{equation} \label{eq:mss_dataset}
  \mathcal{D} = \left\{\left(\tau_i, \delta_i, \left\{\left(x_i(t_{i,j}^{ss}),t_{i,j}^{ss}\right)\right\}_{j=1}^{M_i}  \right)\right\}_{i=1}^N.
\end{equation}

An important difference compared to the conventional case is that the data contains $M_i$ snapshots from subject $i$, and therefore $M_i$ different predictions can be made, one for each time $t_{i,j}^{ss}$. This is especially important when it comes to determining the likelihood.

\subsection{Likelihood}
In the case of a single snapshot $\left(\tau_i, \delta_i, \left(x(t_{i}^{ss}),t_{i}^{ss} \right)\right)$ from individual $i$, the likelihood can be defined similarly as before as
\begin{equation} \label{eq:lik_S_ss}
  L\left(\theta \mid \tau_i, \delta_i, \left(x(t_{i}^{ss}),t_{i}^{ss} \right)\right) = 
  \begin{cases}
    f_\theta(\tau_i;t_{i}^{ss}, x_{i}), & \delta=1\\
    S_\theta(\tau_i;t_{i}^{ss}, x_{i}),& \delta =0
  \end{cases}.
\end{equation}
Based on this, it is tempting to assume that the likelihood in the case when there is more than one snapshot from an individual can be defined as the product of the likelihood based on a single snapshot as
\begin{multline}
  \tilde{L}\left(\theta \mid \tau_i, \delta_i, \left\{ \left(x(t_{i,j}^{ss}),t_{i,j}^{ss} \right) \right\}_{j=1}^{M_i} \right) 
  \\= 
  \prod_{j=1}^{M_i}  L\left(\theta \mid \tau_i, \delta_i, \left(x(t_{i,j}^{ss}),t_{i,j}^{ss} \right)\right),
\end{multline}
and would be motivated if the data in each factor were independent. 
However, this is clearly not the case since all contain $\tau_i$ and $\delta_i$, and in most cases, there are also dependencies between the different snapshots. This means that the usual motivation for the likelihood to factor does not hold, and can not be used to show that $\tilde{L}$ is a proper likelihood. In fact, there is nothing that suggests that $\tilde{L}$ is a proper likelihood, and as will be seen in \secref{sec:simulated_example}, it is easy to find cases where it is not. At the same time, by assuming that it can be used as a proper likelihood it is, as will be shown later, possible to extend the maximum-likelihood training described in \secref{sec:conv_training} to this case in a straightforward way. We therefore call $\tilde{L}$ a quasi-likelihood and in \secref{sec:ss_strategies} we will investigate under what circumstances it can substitute as a likelihood. 

Since observations from two different individuals are assumed independent, together with the assumption that the quasi-likelihood behaves like a proper likelihood, it follows naturally that the quasi-likelihood considering the complete dataset 
\eqref{eq:mss_dataset} of $N$ individuals factors into
\begin{equation}\label{eq:quasi_likelhood}
  \tilde{L}\left(\theta \mid \mathcal{D}\right) = 
  \prod_{i=1}^{N}
  \tilde{L}\left(\theta \mid \tau_i, \delta_i, \left\{ \left(x(t_{i,j}^{ss}),t_{i,j}^{ss} \right) \right\}_{j=1}^{M_i} \right).
\end{equation}

\section{Maximum Quasi-Likelihood Training}\label{sec:ss_strategies}
By comparing the likelihood of a single snapshot \eqref{eq:lik_S_ss} with the likelihood in the conventional case \eqref{eq:conv_loglik}, it can be seen that they are essentially the same if we in the snapshot case consider the pair $\left(x_i(t_{i,j}), t_{i,j}\right)$ as the covariate $x$ in the conventional case. This means that, by considering each snapshot from all individuals as a single observation, corresponds to the dataset 
\begin{equation}\label{eq:quasi_dataset}
  \begin{split}
  \tilde{\mathcal{D}} = \cup_{i=1}^N \left\{ \left(\tau_i, \delta_i, \left(x(t_{i,j}^{ss}),t_{i,j}^{ss} \right)\right) \right\}_{j=1}^{M_i}
                      \\= 
                      \begin{Bmatrix}
                        \left(\tau_1, \delta_1, \left(x_1(t_{1,1}^{ss}),t_{1,1}^{ss} \right)\right)\\
                        \vdots\\
                        \left(\tau_1, \delta_1, \left(x_1(t_{1,M_1}^{ss}),t_{1,M_1}^{ss} \right)\right)\\
                        \left(\tau_2, \delta_2, \left(x_2(t_{2,1}^{ss}),t_{2,1}^{ss} \right)\right)\\
                        \vdots\\
                        \left(\tau_N, \delta_N, \left(x_N(t_{N,M_N}^{ss}),t_{N,M_N}^{ss} \right)\right)
                      \end{Bmatrix}
                    \end{split},
\end{equation} 
maximizing the quasi-likelihood \eqref{eq:quasi_likelhood} for the dataset $\mathcal{D}$ is the same as maximizing the conventional likelihood \eqref{eq:conv_lik} for the dataset $\tilde{\mathcal{D}}$. Consequently, a usage-specific survival model can be trained using maximum-likelihood training described in \secref{sec:conv_training} by simply using the dataset $\tilde{\mathcal{D}}$.

This means that in principle any method for training conventional survival models can be used for training usage-specific survival models. However, it remains to show under what circumstances maximizing the quasi-likelihood gives the desired result. 

\subsection{An Example Where the Quasi-Likelihood is Inappropriate}\label{sec:illustrative_example}
We will here give an illustrative example when maximizing the quasi-likelihood fails to give a desirable result. 

\begin{figure}[h]
  \centering
  \includegraphics[width=\columnwidth]{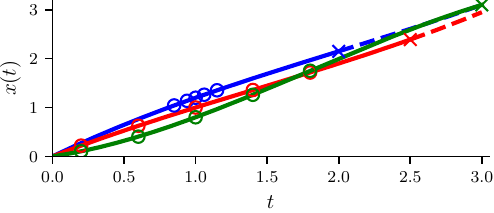}
  \caption{An illustrative example showing the accumulative usage $x(t)$ for three individual. Included are also their respective failure times marked by crosses, and times from which snapshots are available marked with circles.
  }
  \label{fig:illustrative_example}
\end{figure}

In \figref{fig:illustrative_example} an illustrative example of data from three individuals is shown. As can be seen, the accumulative usage is quite similar which indicates that the individuals are used similarly, and therefore predictions about their failure times should be similar. For example, when considering predictions based on usage up to age $t=1$, the accumulative usage is around 1 for all individuals, and since the failure times are 2, 2.5, and 3 a reasonable conclusion would be that a failure sometime between 2 and 3 time-units is likely for an individual that is used in this way. 

However, when looking at the snapshots, it can be seen that the blue individual has 5 snapshots around age $t=1$, while the others only have 1. This means that if a model is trained using this data, the failure time of the blue individual will be overrepresented and most likely the resulting model will predict failure times closer to 2 time units.

Since all individuals have the same number of snapshots in total, it can also be concluded that the problem in this example is how the snapshot times are distributed and not how many there are from each individual. 

\subsection{Homogeneously Sampled Datasets}
A conclusion that can be made from the previous example is that for any time $t_0 \geq 0$ the number of snapshots from a time close to $t_0$, say between $t_0-\Delta t$ and $t_0+\Delta  t$ for some $\Delta t$,  should be similar for all individuals, otherwise the predictions will be biased towards individuals with more snapshots around $t_0$. 

By taking the argument above to the extreme, and saying that for any $t_0\geq 0$ and $\Delta t>0$ the number of snapshots between $t_0-\Delta t$ and $t_0+\Delta  t$ should be the same from all individuals we get the following requirement: if individual $i$ has a snapshot from time $t$ then so must individual $j$, since otherwise, subject $i$ would have more snapshots around time $t$ for some $\Delta t>0$. It follows that the data from any individual must contain a snapshot from time $t$ and that the data must be on the form 
\begin{equation} \label{eq:mss_dataset}
  \mathcal{D} = \left\{\left(\tau_i, \delta_i, \left\{\left(x_i(t_{k}^{ss}),t_{k}^{ss}\right)\right\}_{k=1}^{M}  \right)\right\}_{i=1}^N
\end{equation}
for some sequence of snapshot times $(t_k^{ss})_{k=1}^M$. 

As discussed in \secref{sec:us_surv_func}, it is often not possible or desirable to have snapshots from times after the individual has failed, and therefore it might be necessary to relax the condition slightly and say that the number of snapshots from all individuals that have not failed before time $t_0+\Delta  t$ should have the same number of snapshots between $t_0-\Delta t$ and $t_0+\Delta  t$. This results in that all individuals do not need to have a snapshot from all $t_k$ for $k=1,2,\ldots,M$, but it is enough that they have snapshots from all times smaller than their failure time, i.e. for all $t_k^{ss}<\tau_i$.  

Based on the discussions above we define the following. 
\begin{definition}[homogeneously sampled]
A dataset \eqref{eq:mss_dataset} is \textit{homogeneously sampled} if there is a sequence $t_1^{ss}<t_2^{ss}<\cdots<t_M^{ss}$ of snapshot times such that
\begin{equation}\label{eq:homogeneously_sampled}
  \mathcal{D} = \left\{\left(\tau_i, \delta_i, \left\{\left(x_i(t_{k}^{ss}),t_{k}^{ss}\right)\right\}_{k=1}^{M_i}  \right)\right\}_{i=1}^N
\end{equation}
where either all $M_i$ are the same (and equal to $M$) or $M_i = \max \left\{ k\in\{1,2,\ldots,M\}: t_k^{ss} \leq \tau_i\right\}$.
\end{definition}
This definition aims to specify when a dataset is suitable for maximum quasi-likelihood training; with the motivation that if it is not, the problems discussed in \secref{sec:illustrative_example} might arise. 

\subsection{Homogeneous Resampling of a Dataset}
If a dataset is not homogeneously sampled it is natural to try to transform it into a homogeneously sampled dataset, and one way to do this is to simply resample the dataset as described below. 

When the snapshots are based on the accumulative data \eqref{eq:accumulative_ss}, $x(t)$ is a continuous function and by interpolating the snapshots from individual $i$ an estimate $\hat{x}_i(t)$ of $x_i(t)$ is obtained for all times $t$ between the first and last snapshot. While the type of interpolation scheme to use probably depends on the situation, we have in this work used linear interpolation.

By defining a sampling grid $G=\left\{g_1,g_2,\ldots,g_M\right\}$, and resampling each individual on $G$, a dataset that is by design homogeneously sampled is obtained. However, it is not always possible to resample all individuals on each point in $G$ since interpolation can only be done between the first and last snapshot, and $G$ might contain points outside this interval. Instead, by letting $I_i$ be the index of all sampling points $g_k\in G$ on which individual $i$ can be resampled on, the resampled dataset can be written
\begin{equation}\label{eq:resampled_dataset}
  \hat{\mathcal{D}}(G) = \left\{\left(\tau_i, \delta_i, \left\{\left(\hat{x}_i(g_{k}),g_{k}\right)\right\}_{k\in I_i  }  \right)\right\}_{i=1}^N.
\end{equation}
By comparing with \eqref{eq:homogeneously_sampled} we see that if, for all individuals, $I_i=\{1,2,\ldots,M_i\}$ with $M_i$ as in \eqref{eq:homogeneously_sampled}, then $\hat{\mathcal{D}}(G)$ is homogeneously sampled.

The condition that $I_i=\{1,2,\ldots,M_i\}$ has two implications: The first is that all individuals must have a snapshot from a time before the first sampling point in $G$, so that $1 \in I_i$. The second is that all individuals have a snapshot from a time later than either the last sampling point in $G$ (corresponding to all $M_i$ the same and equal to $M$), or the last sampling point in $G$ that is smaller than the failure time of the individual (corresponding to when $M_i$ depends on the failure time in \eqref{eq:homogeneously_sampled}). In practice, however, we have found that it is often enough that most individuals can be sampled uniformly on the selected grid.

\subsection{Epochwise Random Resampling}

Consider a grid $G$ of size $M$ points. If $M$ is small, it is likely that the resulting model will only be accurate for predictions based on snapshots from times around the points in $G$, which can be interpreted as a type of overfitting to the points in $G$. On the other hand, if $M$ is large, the dataset will become large making the training slow. A compromise must therefore be made when choosing $G$.

A way to circumvent this compromise is to use a smaller grid, but in each epoch change the grid. That is, in epoch $n$ use the grid 
\begin{equation}
  G_n=\left\{g_1^{(n)},g_2^{(n)},\ldots,g_M^{(n)}\right\}.
\end{equation} 
In this way, by using different $G_n$ in each epoch, the resulting model will not be overfitted for a particular grid; at the same time, the size of the dataset will be kept small.

Generating $G_n$ can of course be done in many ways. To train a model that can be used for predictions based on snapshots from times between $t_{min}^{ss}$ and $t_{max}^{ss}$, we propose to generate the grid points as
\begin{equation}\label{eq:unifrom_random_resampling}
  g_k^{(n)} = t_{min}^{ss} + \frac{t_{max}^{ss}-t_{min}^{ss}}{M} u_k^{(n)}
\end{equation} 
where $u_k^{(n)}$ are independent and uniformly distributed between zero and one.

\subsection{Training for Remaining-Life Predictions}\label{sec:remaining_life_modifications}
So far we have only discussed how to train a model to predict the total life of the component, and not to predict the remaining life. However, this is simply done by realizing that when considering the model $S^r$ in \eqref{eq:rem_ub_surv} for the remaining life, the likelihood  \ref{eq:lik_S_ss} for a single snapshot becomes  
\begin{equation} \label{eq:lik_remaining_S_ss_remaining}
  L\left(\theta \mid \tau_i, \delta_i, \left(x(t_{i}),t_{i} \right)\right) = 
  \begin{cases}
    f^r_\theta(\tau_i-t_{i};t_{i}, x_{i}), & \delta=1\\
    S^r_\theta(\tau_i-t_{i};t_{i}, x_{i}),& \delta =0
  \end{cases}.
\end{equation}
where $\tau_i-t_{i}$ is the remaining life at the time of the snapshot. This means that by simply exchanging $\tau_i$ for $\tau_i-t_{i,j}$ when creating the dataset \eqref{eq:quasi_dataset} and applying maximum quasi-likelihood training, a model for the remaining life is trained.

\section{A Simulated Example}\label{sec:simulated_example}

In this section, a simulated dataset is used as an example where the properties of maximum quasi-likelihood training are investigated.

We consider a system whose usage $U$ is constant over time, but varies among individuals as $U\sim \text{Uniform}(1,5)$. For a specific usage $U=u$, the failure times are modeled using a Weibull distribution with shape parameter $k=2$ and $u$ as scale parameter, giving the  survival function
\begin{equation}\label{eq:simiulated_surv}
  S(t; u) = P(T>t\mid U = u) =e^{-\left({ut}\right)^2}
\end{equation}
To model the operational data for a given usage $u$, the accumulative usage 
\begin{equation}
  x(t) = \int_{0}^{t} u \, d\tau = ut
\end{equation}
is used. This means that the operational data of the population is described by 
\begin{equation}
  X(t) = Ut \sim \text{Uniform}(t,5t).
\end{equation}

\subsection{Dataset Generation}
To simulate a population of $N$ individuals, for each $i$ a usage $u_i$ is generated from $\text{Uniform}(1,5)$ and a failure time is then generated from the distribution in \eqref{eq:simiulated_surv}. Right-censoring is introduced by generating a censoring time from a Uniform$(0,3)$ distribution, and if the failure time is larger than the censoring time the individual is censored. Since $x_i(t)$ is linear in $t$ for all individuals, only one snapshot from the time of failure is needed since $x_i(t)$ can be reconstructed from this when the dataset is resampled.

\subsection{True Usage-Specific Survival Function}
A benefit of this example is that the true usage-specific survival
function can be determined. For $t_0>0$ it becomes
\begin{equation}
  \begin{split}
  S(t; x,t_0) &=  P\left(T>t\mid X(t_0) = x\right) = P\left(T>t\mid t_0 U = x\right)\\
              &= P\left(T>t\mid  U = \frac{x}{t_0}\right) = S\left(t; \frac{x}{t_0}\right)
              = e^{-\left(t\frac{x}{t_0}\right)^2}
  \end{split}
\end{equation}
and 
\begin{equation}
  S^r(t; x,t_0) = \frac{S(t+t_0; x,t_0)}{S(t_0; x,t_0)} 
  = e^{-\left(t\frac{x}{t_0}\right)^2+ x^2}.
\end{equation}
For $t_0 = 0$, on the other hand, $x_i(t_0)=0$ and no information about the individual is provided, instead, we have 
\begin{equation}
  \begin{split}
  S(t; x=0, t_0=0) &=  P\left(T>t\mid X(0) = 0\right) \\
              &=\int_{1}^5 \frac{1}{5-1} P\left(T>t\mid  U = u\right)\, du
  \end{split}
\end{equation}
i.e. the population average (which does not have a convenient expression). 
This also means that there is a discontinuity in $S(t; x, t_0)$ at $t_0=0$.

\subsection{Resampling the Dataset}
The fact that there is a discontinuity in $S(t; x, t_0)$ at $t_0=0$ makes it difficult to train a network to produce accurate predictions around $t_0=0$. For this reason, a lower limit $t_{min}^{ss}=0.1$ is put on the prediction time $t_0$. Since most individuals tend to fail before one time unit an upper limit $t_{max}^{ss}=1$ is also used. This means that only snapshot times between these two limits need to be considered.

Both fixed resampling and epochwise resampling were used to train models. For fixed resampling an equidistant grid between $t_{min}^{ss}$ and $t_{max}^{ss}$ was used and for epochwise random resampling the sampling in \eqref{eq:unifrom_random_resampling} was used with $t_{min}^{ss}$ and $t_{max}^{ss}$ as the limits.

\subsection{Training and Network Architecture}  \label{sec:sim_training}

During training, the dataset was first split into two separate datasets, one for training and one for validation; 85~\% of the individuals were used for training and the rest for validation, note that the split was done based on individuals so that snapshots from the same individual is not present in both the training and validation data. Each model was trained for 200 epochs using the Adam optimizer; however, the state of the model on the epoch with the lowest validation loss was used in the end. 

It was found that a small network with only two layers of 32 neurons each is sufficiently expressible for this type of data. Dropout was also evaluated, but it was found that it did not improve the result. 15 different learning rates between $10^{-2}$ and $0.25$ following a geometric progression were evaluated. Since there are stochastic elements in the training 40 different randomly generated representations of each dataset were evaluated and the learning rate that performed best on average was selected.

To evaluate the models a separate dataset for testing of 500 individuals was utilized. All 40 models for each representation of the dataset were evaluated and the mean of the loss was noted.

\subsection{Results}

\begin{figure}[h]
  \centering
  \includegraphics[width=\columnwidth]{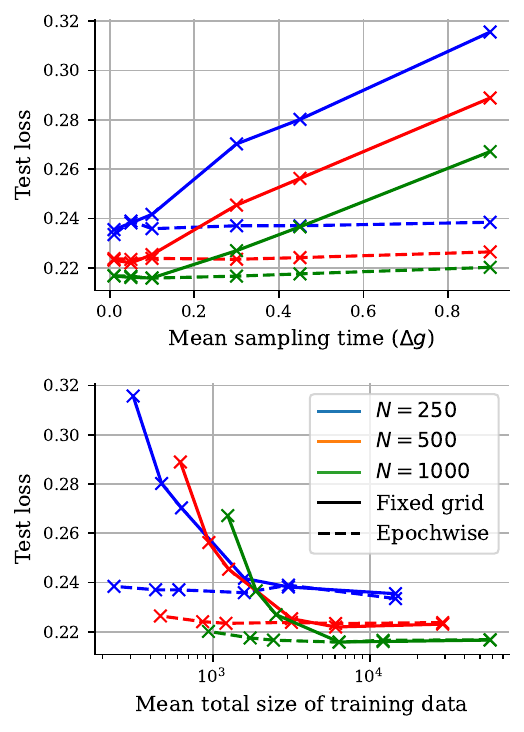}
  \caption{
    Results from models trained on datasets of three different sizes, for different numbers of samples in the resampling, and for both fixed sampling and epochwise random resampling. The results are shown both as a function of the sampling time (distance between two samples in the sampling grid), and total size of the training data; since for epochwise resampling, the grid varies the mean is used in both cases. The test loss is the mean loss for the 40 models evaluated on the test set as described in \secref{sec:sim_training}}
  \label{fig:simulated_example_1}
\end{figure}

\figref{fig:simulated_example_1} the result from training models on datasets with a varying number of individuals, different numbers of samples in the resampling, and for both fixed sampling and epochwise random resampling. Here it can be seen that increasing the number of samples (reducing the sampling time) improves the result, at least until some point. This holds both for fixed resampling and epochwise resampling; however, epochwise resampling is less sensitive to the number of samples and always performs better than fixed resampling for the same amount of data. It can also be seen that increasing the number of individuals improves the result which indicates consistency.

\begin{figure}
  \centering
  \includegraphics[width=\columnwidth]{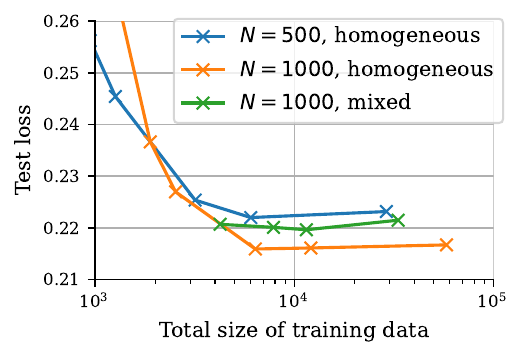}
  \caption{
    Results from models trained using homogeneously sampled datasets as well as the mixed dataset containing two different sampling densities.}
  \label{fig:simulated_example_2}
\end{figure}

In \figref{fig:simulated_example_1} all data is homogeneously sampled and, as discussed above, the results are as expected. In \figref{fig:simulated_example_2} an example of models trained on data that is not homogeneously sampled is shown. The green line in this figure comes from models trained on a dataset consisting of the union of two datasets both of size 500 individuals and equidistantly sampled, but one of them has been sampled 10 times more densely. As can be seen, the performance of the models trained on the mixed datasets is closer to the performance of the models trained on the homogeneously sampled data from 500 individuals, even though the total number of individuals in this data is 1000 individuals. A logical explanation for this is that since the 500 individuals that are sampled with higher frequency are over-represented in the data, the predictions from the resulting model will be similar to a model trained on the data from these 500 individuals. 

\section{Starter Battery Failure Data}\label{sec:battery_data}
In this section, an experimental dataset consisting of starter battery failure times, from a fleet of around 25,000 vehicles, is used to evaluate the methodology on experimental data. 

\subsection{Description of the Dataset}
The dataset has a censoring rate of 74~\%, meaning that a failure was only observed for 26~\% of the vehicles. The snapshots in this data set consist of operational data that is aggregated over time in the vehicles' control units and stored during specific times during the vehicles' lifetime, for example when the vehicle visits a workshop. The frequency of which data is stored for a specific vehicle is fairly constant over time, but varies greatly from vehicle to vehicle; from a few snapshots per year to one snapshot per week. 

The information in each snapshot is based on various signals available in the vehicles' control system selected by experts to be informative for predicting battery failures; for example, mileage, engine load, and number of engine starts. Most of the signals are stored as histograms indicating the amount of time the signal has spent in different intervals. For more information about the dataset see \cite{holmer2023energy,voronovMachineLearningModels2020}.

\subsection{Training}
The dataset was first split into three parts in the conventional way: 70~\% for training, 15~\% for validation, and 15~\% for testing. To make sure that data from a specific vehicle only ends up in one of the datasets the split was done based on the number of vehicles in the datasets.

The same training as described in \secref{sec:sim_training} was used for this dataset as well, but with a hyperparameter search over learning rate, number of nodes per layer, and amount of dropout, based on the search spaces in \tabref{tab:search_space}.
\begin{table}[h!]
  \centering
  \begin{tabular}{l | r } 
   \hline
   Parameter & Search space \\ [0.5ex] 
   \hline\hline
   Learning rate & $\left\{0.001     , 0.00215, 0.00464, 0.01      , 0.0215,
   0.0464, 0.1   \right\}$ \\
   Nodes per layer &  $\left\{ 64, 128, 256 \right\}$  \\ 
   Dropout & $\left\{ 0~\%,10~\%,25~\%,50~\%  \right\}$ 
   \\[1ex] 
   \hline
  \end{tabular}
  \caption{Search spaces used in the hyperparameter search.}
  \label{tab:search_space}
  \end{table}

In total 7 different models were trained. The first two models are models trained for a specific prediction time of one respective two years, which are trained using a single snapshot from that time from each vehicle; this means that the training data for these models are independent and the conventional maximum likelihood training can be used. The next three models are all based on resampled versions of the dataset with varying numbers of samples in the resampling; all were equidistantly sampled between 0.5 and 2.5 years, and the number of samples was 4, 5, and 12. The last model was trained using the original dataset without any resampling, but only using snapshots from times between 0.5 and 2.5 years. 

\subsection{Evaluation}
To evaluate the models three different resampled versions of the test set were evaluated using two different metrics. 

Two of the test sets were created by resampling the original test set to only include data from one year and two years, respectively; and a third dataset was created by resampling the dataset using 15 random samples based on \eqref{eq:unifrom_random_resampling}.

As an evaluation metric the quasi-log-likelihood was used, which for the test sets resampled on a single time becomes a proper likelihood. As an additional evaluation metric, the Brier score was used, which is defined for the dataset \eqref{eq:quasi_dataset} as
\begin{multline}
  J_{BS}(t) = \frac{1}{\sum_{i=1}^{M_i} M_i} \sum_{i=1}^{N} \sum_{j=1}^{M_i}
  \left(
  \frac{\mathbb{I}_{ \left\{ \tau_i>t \right\} \left(1-S(t;t_{i,j}^{ss}, x_i(t_{i,j}^{ss})) \right)^2 }}{G(t)} \right.
  \\\left.+
  \frac{\mathbb{I}_{ \left\{ \tau_i>t,\delta_i=1 \right\} S(t;t_{i,j}^{ss}, x_i(t_{i,j}^{ss}))^2}}{G(\tau_i)} \right)
\end{multline}
where $G$ is the Kapplan-Meier estimate of the censoring distribution. Since the Brier score is a function of time, the results are presented in terms of the integrated Brier score defined as 
\begin{equation}
  J_{IBS} = \frac{1}{\max_{i} \tau_{i}} \int_0^{\max_{i} \tau_{i}} J_{fBS}(t)
\end{equation}
which is evaluated numerically.
\subsection{Results}
This section contains possibly sensitive data and will be made available in the final version.

\section{Conclusion}
This paper proposes a methodology for defining and training data-driven usage-specific survival models based on continuously gathered operational data. The models can be used to predict the remaining life of a component based on its usage history and as such fit well into many predictive maintenance applications. 

The methodology is based on conventional neural network-based survival models that are trained using data that is continuously gathered and stored at specific times, called snapshots. The fact that the data can contain more than one snapshot from a specific individual means that the standard maximum likelihood training can not be directly applied since the data is not independent. However, the papers show that if the data is in a specific format where all snapshot times are from the same time for all individuals, called homogeneously sampled, maximum likelihood training can be applied and produce desirable results.  

In many cases, the data is not homogeneously sampled and in this case, it is proposed to resample the data to make it homogeneously sampled. The results from applying this to a dataset of starter battery failures indicate that this is a promising approach. However, the results also show that the number of samples that the dataset is resampled with is an important parameter; it should be chosen large enough to produce good results, but this also increases the size of the dataset which makes training slow. To reduce the number of samples needed to produce good results it is also proposed that, instead of resampling the dataset once before the training starts, randomly resample the dataset at the start of each epoch during the training. Randomly resampling the dataset on each epoch is shown to greatly reduce the number of samples needed to produce the same results as that obtained from only resampling the dataset once. 

\bibliography{bibliography_BBT}

\end{document}